\documentclass{article}
\usepackage{spconf,amsmath,graphicx}
\usepackage{color} 
\usepackage{amsfonts} 
\usepackage{hyperref} 
\usepackage{cite}     
\usepackage{subfig}   
\usepackage{url}

\title{Towards Structured Deep Neural Network \\ for Automatic Speech Recognition}
\name{{\em Yi-Hsiu Liao$^1$, Hung-yi Lee$^2$, Lin-shan Lee$^3$}}
\vspace{-8mm}
\address{
  National Taiwan University\\
  {\small \tt r03921048@ntu.edu.tw$^1$, hungyilee@ntu.edu.tw$^2$, lslee@gate.sinica.edu.tw$^3$}
}
\begin{document}
%
\maketitle
%
\begin{abstract}
  In this paper we propose the Structured Deep Neural Network (structured DNN) as a structured and deep learning framework. This approach can learn to find the best structured object (such as a label sequence) given a structured input (such as a vector sequence) by globally considering the mapping relationships between the structures rather than item by item. 
  When automatic speech recognition is viewed as a special case of such a structured learning problem, where we have the acoustic vector sequence as the input and the phoneme label sequence as the output, it becomes possible to comprehensively learn utterance by utterance as a whole, rather than frame by frame. 
  Structured Support Vector Machine (structured SVM) was proposed to perform ASR with structured learning previously, but limited by the linear nature of SVM. Here we propose structured DNN to use nonlinear transformations in multi-layers as a structured and deep learning approach. This approach was shown to beat structured SVM in preliminary experiments on TIMIT.

\end{abstract}
\begin{keywords}
structured learning, deep neural network
\end{keywords}
%


\vspace{-3mm}
\section{Introduction}
\vspace{-2mm}
\label{sec:intro}

With the maturity of machine learning, great efforts have been made to try to integrate more machine learning concepts into the Hidden Markov Model (HMM).
Using Deep Neural Networks (DNN)\cite{hinton2006fast, hinton2006reducing} with HMM is a good example\cite{dahl2012context, mohamed2012acoustic, hinton2012deep}.
In general, HMMs consider the phoneme structure by states and the transitions among them, but trained primarily on frame level regardless of being based on DNN\cite{tuske2014acoustic, vinyals2013deep} or Gaussian Mixture Model (or subspace GMM, SGMM\cite{povey2010subspace}). 
Under HMM framework\cite{rabiner1993fundamentals}, the hierarchical structure of an utterance is taken care of by the HMM and their states, the lexicon and the language model, which are respectively learned separetely from disjoint sets of knowledge sources. On the other hand, it is well known that there may exist some underlying overall structures for the utterances behind the signals which may be helpful to recognition. If we can learn such structures comprehensively from the signals of the entire utterance globally, the recognition scenario may be different.

On the contrary, structured learning has been substantially investigated in machine learning, which tries to learn the complicated structures exhibited by the data.
Conditional Random Fields (CRF)\cite{lafferty2001conditional, mccallum2003early, gunawardana2005hidden, zweig2009segmental, sung2009hidden, yu2010deep} and structured Support Vector Machine (SVM)\cite{tsochantaridis2005large, zhang2011structured, zhang2013structured} are good example approaches. Recently, structured SVM has been used to perform initial phoneme recognition by learning the relationships between the acoustic vector sequence and the phoneme label sequence of the whole utterance jointly rather than on the frame level or from different sets of knowledge sources\cite{tang2010initial}, utilizing the nice properties of SVM\cite{boser1992training} to classify the structured patterns of utterances with maximized margin. 
However, both CRF and structured SVM are linear, therefore limited in analyzing speech signals. 

In this paper, we extend the above structured SVM approach to phoneme recognition using a structured DNN including nonlinear units in multi-layers, but similarly learning the global mapping relationships from an acoustic vector sequence to a phoneme label sequence for a whole utterance. 
In recent work, the front-end feature extraction DNN has been integrated with SVM~\cite{zhangdeep} and Weighted Finite-State Transducers (WFST)~\cite{kubo2012integrating}, but here we further integrate the front-end DNN with structured DNN, which is completely different from the previous work.


\vspace{-3mm}
\section{Proposed Approach -- Structured Deep Neural Network}
\begin{figure}[htb]
  \centering
  \includegraphics[width=\linewidth]{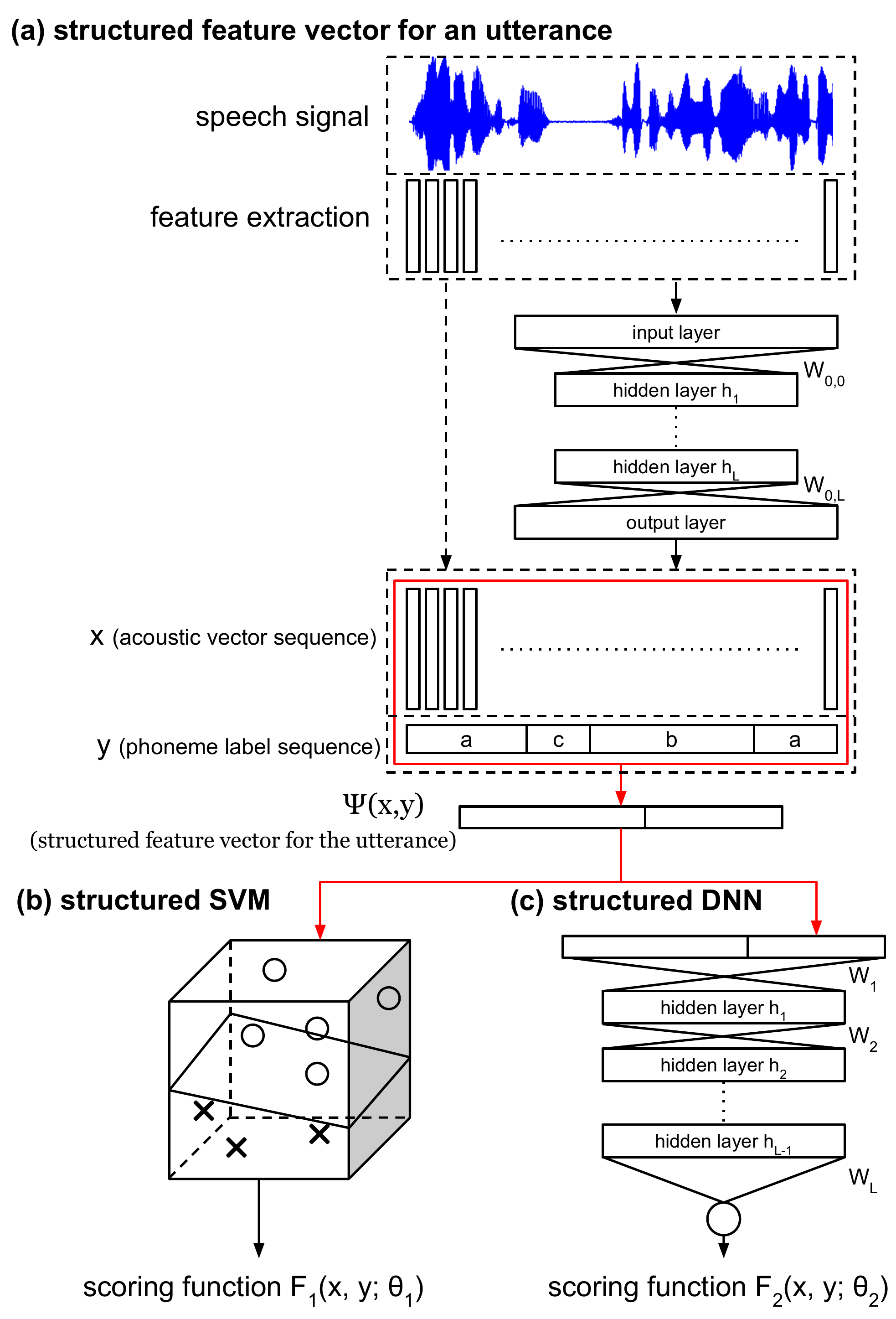}
  \vspace{-5mm}
  \caption{The concept of Structured SVM and Structured Deep Neural Network: (a) the structured feature vector $\Psi(\mathbf{x}, \mathbf{y})$ for an utterance, (b) structured SVM and (c) structured DNN. } 
  \vspace{-5mm}
  \label{fig:SDNN}
\end{figure}

\vspace{-2mm}
\label{sec:arch}
The whole picture of the concept of the structured DNN for phoneme recognition is in Fig.~\ref{fig:SDNN}. Given an utterance with an acoustic vector sequence $\mathbf{x}$ and a corresponding phoneme label sequence $\mathbf{y}$, we can first obtain a structured feature vector $\Psi(\mathbf{x}, \mathbf{y})$ representing $\mathbf{x}$ and $\mathbf{y}$ and the relationships between them as in Fig.~\ref{fig:SDNN}(a) (details of $\Psi(\mathbf{x}, \mathbf{y})$ are given in Section \ref{sec:struct}), and then feed it into either an SVM as in Fig.~\ref{fig:SDNN}(b) or a DNN as in Fig.~\ref{fig:SDNN}(c) to get a score by a scoring function $F_1(\mathbf{x}, \mathbf{y}; \theta_1)$ or $F_2(\mathbf{x}, \mathbf{y}; \theta_2)$, where $\theta_1$ and $\theta_2$ are the parameter sets for the SVM and DNN respectively.
Here the acoustic vector sequence $\mathbf{x}$ can be raw acoustic features like filter bank outputs or phoneme posteriorgram vectors generated from the DNN in Fig.~\ref{fig:SDNN}(a).
Because both $\mathbf{x}$ and $\mathbf{y}$ represent the entire utterance by a structure (sequence), and either SVM or DNN learns to map the pair of $(\mathbf{x}, \mathbf{y})$ to a score on the utterance level globally rather than on the frame level, this is structured learning optimized on the utterance level.

\vspace{-5mm}
\subsection{Structured Learning Concept}
\vspace{-1mm}
In structured learning, both the desired outputs $\mathbf{y}$ and the input objects $\mathbf{x}$ can be sequences,  trees, lattices, or graphs, rather than simply classes or real numbers.
In the context of supervised learning for phoneme recognition for utterances, we are given a set of training utterances, $(\mathbf{x}_1, \mathbf{y}_1) , ... , (\mathbf{x}_N, \mathbf{y}_N) \in \mathbf{X} \times \mathbf{Y} $, where $\mathbf{x}_i$ is the acoustic vector sequence of the i-th utterance, $\mathbf{y}_i$ the corresponding reference phoneme label sequence, and we wish to assign correct phoneme label sequences to unknown utterance. 

We first define a function $f(\mathbf{x}; \theta) =\mathbf{y} : \mathbf{X} \rightarrow \mathbf{Y}$ , mapping each acoustic vector sequence $\mathbf{x}$ to a phoneme label sequence $\mathbf{y}$, where $\theta$ is the parameter set be learned. One way to achieve this is to assign every possible phoneme label sequence $\mathbf{y}$ given an acoustic vector sequence $\mathbf{x}$ a score by a scoring function $F(\mathbf{x}, \mathbf{y}; \theta) : \mathbf{X} \times \mathbf{Y} \rightarrow \mathbb{R}$, and take the phoneme label sequence $\mathbf{y}$ giving the highest score as the output of $f(\mathbf{x}; \theta)$,
\vspace{-3mm}
\begin{equation}
  f(\mathbf{x}; \theta) = \arg\max_{\mathbf{y}\in \mathbf{Y}} F(\mathbf{x}, \mathbf{y}; \theta).
  \label{equ:sl}
\end{equation}

\begin{figure}[htb]
  \centering
  \includegraphics[width=60mm]{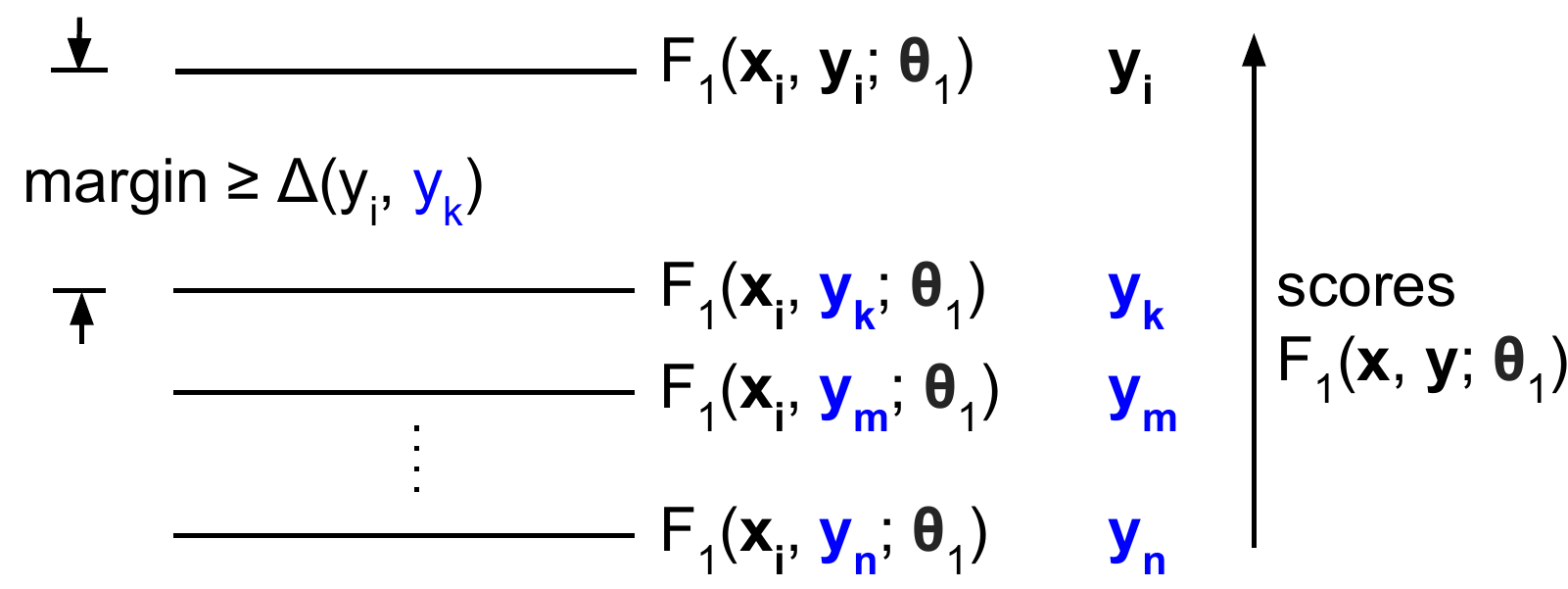} 
  \caption{{We need to maximize the margin between the correct label sequence ($y_i$) and all the other incorrect label sequences (in blue color). }}
\vspace{-5mm}
  \label{fig:margin}
\end{figure}

\vspace{-5mm}
\subsection{Structured SVM}
\label{subsec:ssvm}

Based on the maximized margin concept of SVM, we wish to maximize not only the score of the correct label sequence, but the margin between the score of the correct label sequence and those of the nearest incorrect label sequences as shown in Figure~\ref{fig:margin}. 
In Figure~\ref{fig:margin}, $y_i$ is the correct label sequence, and all the other incorrect label sequences are in blue.
The score of the correct label sequence $F_1(\mathbf{x}, \mathbf{y}_i; \theta_1)$ is higher than the highest score among the incorrect label sequences, $F_1(\mathbf{x}, \mathbf{y}_{k}; \theta_1)$, by $\Delta( \mathbf{y}_i, \mathbf{y}_{k} )$, which is the difference between the scores of the two sequences $\mathbf{y}_{i}$ and $\mathbf{y}_{k}$. All incorrect label sequences have scores below that of the correct sequence by at least $\Delta(\mathbf{y}_i, \mathbf{y}_k)$, or the margin.
Maximizing this margin is the learning target of structured SVM.
The scoring function used in structured SVM is as below, which is linear.
\begin{equation}
  F_1(\mathbf{x}, \mathbf{y}; \theta_1) = \left \langle \theta_1, \Psi(\mathbf{x}, \mathbf{y}) \right \rangle ,
\label{equ:ssvm}
\end{equation}
where $\Psi(\mathbf{x}, \mathbf{y})$ is the structured feature vector mentioned above and shown in Figure \ref{fig:SDNN}, representing the structured relationship between $\mathbf{x}$ and $\mathbf{y}$, $\theta_1$ is in vector form, and $\left \langle \cdot , \cdot \right \rangle $ represents inner product. We can then train the parameter vector $\theta_1$ using training instances $\left \{ (\mathbf{x}_i, \mathbf{y}_i), i = 1, 2,...,\text{N} \right \}$ 
subject to the following formula,
\vspace{-3mm}
\begin{align}
\min_{\theta_1, \xi_i}{\left \| \theta_1 \right \|^2 + C\sum_{i=1}^{N}{\xi_i}},\qquad 
\xi_i > 0, \qquad
\forall{\mathbf{y} \in \mathbf{Y}},& \notag \\
F_1(\mathbf{x_i}, \mathbf{y_i}; \theta_1) - F_1(\mathbf{x_i}, \mathbf{y}; \theta_1) + \xi_i \geq \Delta(\mathbf{y}_i, \mathbf{y}),& 
\label{equ:ssvm_risk}
\end{align}
where $C$ is the cost balancing the model complexity ($\theta_1$) with the inequality, and $\xi_i$ is the slack variable for the inequality, and $\Delta(\cdot, \cdot) : \mathbf{Y} \times \mathbf{Y} \rightarrow \mathbb{R}^+ $ measures the distance between two label sequences. 
Phone error rate is used as the distance in this paper, but other evaluation metrics are also feasible.
Optimizing the function above is equivalent to maximizing the margin separating the scores between  the correct label sequence and all other label sequences for each training sample $\mathbf{x}_i$. 
Formula (\ref{equ:ssvm_risk}) can be solved by quadratic programming and the cutting-plane algorithm\cite{joachims2009cutting}, and is equivalent to the following formula:
\vspace{-3mm}
\begin{multline}
\min_{\theta_1}{\left \| \theta_1 \right \|^2 + C\sum_{i=1}^{N}{L_i(\theta_1)}}, \\
L_i(\theta_1) = \max_{\mathbf{y}} max(0, F_1(\mathbf{x}_i, \mathbf{y}; \theta_1) + \Delta(\mathbf{y}_i, \mathbf{y}) \\
-F_1(\mathbf{x}_i, \mathbf{y}_i; \theta_1)),  
\label{equ:ssvm_loss}
\end{multline}
In (\ref{equ:ssvm_loss}), $L_i(\theta_1)$ is the loss function for each example $(\mathbf{x}_i, \mathbf{y}_i)$, and $max(0, \cdot)$ (the inner max operator) in $L_i(\theta_1)$ is a hinge loss function, which  penalizes the model if the inequality in (\ref{equ:ssvm_risk}) does not hold.
Formula (\ref{equ:ssvm_loss}) is helpful in understanding the concept of the cost function for structured DNN in Subsection~\ref{subsubsec:cost_margin}.
With the scoring function $F_1(\mathbf{x}, \mathbf{y}; \theta_1)$ and the trained parameter set $\theta_1$, we can find the label sequence $\mathbf{y}$ for the acoustic vector sequence $\mathbf{x}$ of any input testing utterance by the well known Viterbi algorithm \cite{joachims2009cutting}.

\vspace{-3mm}
\subsection{Structured Deep Neural Network (Structured DNN)}
The assumption of the linear scoring function as in (\ref{equ:ssvm}) makes structured SVM limited. Instead, the proposed structured DNN uses a series of nonlinear transforms to build the scoring function $F_2(\mathbf{x}, \mathbf{y}; \theta_2)$ with L hidden layers to evaluate a single output value $F_2(\mathbf{x}, \mathbf{y}; \theta_2)$ as in Fig. \ref{fig:SDNN}(c).
\begin{align*}
    \mathbf{h}_1&=\sigma (W_0 \cdot \Psi(\mathbf{x}, \mathbf{y})) &  \\
    \mathbf{h}_l&=\sigma (W_{l-1} \cdot \mathbf{h}_{l-1}),  &2 \leq l \leq L \\
\end{align*}
\vspace{-8mm}
\begin{equation}
  F_2(\mathbf{x}, \mathbf{y}; \theta_2) = \sigma (W_{L} \cdot \mathbf{h}_{L}),
\label{equ:sdnn}
\end{equation}
where $W_i$ is weight matrix (including the bias) of layer i, $\sigma(\cdot)$ a nonlinear transform (sigmoid is used here), $h_i$ the output vector of hidden layer i, and the set of all DNN parameters ($W_0$, $W_1$, $W_2$,..., $W_L$) is $\theta_2$. Note that the last weight matrix $W_L$ is a vector, because this DNN gives only a single value as the output.
Two different lost functions for learning structured DNN are defined in this work and described respectively in Subsections~\ref{subsubsec:cost_mse} and~\ref{subsubsec:cost_margin}.

\vspace{-3mm}
\subsubsection{Approximating phoneme accuracy (Approx. Ph. Acc)}\label{subsubsec:cost_mse}


First, the label phoneme accuracy for a label sequence $\mathbf{y}$ is defined as $C(\mathbf{y}_i, \mathbf{y}) = 1 - \Delta(\mathbf{y}_i, \mathbf{y})$, where $\mathbf{y}_i$ is the correct label sequence, and $\Delta(\mathbf{y}_i, \mathbf{y})$ is the phoneme error rate of $\mathbf{y}$ given the correct label sequence $\mathbf{y}_i$.
The parameter set $\theta_2$ of structured DNN can be trained by minimizing the following loss function,
\begin{equation}
  L(\theta_2) = \sum_{i=1}^{N} \sum_{\mathbf{\mathbf{y} \in \mathbf{Y}}}{   \big[C(\mathbf{y}_i, \mathbf{y}) - F_2(\mathbf{x_i}, \mathbf{y}; \theta_2) \big]^2 }.
  \label{equ:loss}
\end{equation}
By minimizing (\ref{equ:loss}), the DNN learns to minimize the mean square error between its output $F_2(\mathbf{x_i}, \mathbf{y}; \theta_2)$ given $\mathbf{x_i}$ and $\mathbf{y}$ and the phoneme accuracy of $\mathbf{y}$, $C(\mathbf{y}_i, \mathbf{y})$, over all training utterances and  for each utterance all possible phoneme sequences $\mathbf{y} \in \mathbf{Y}$.
In other words, the score function $F_2(\mathbf{x}, \mathbf{y}; \theta_2)$ thus learned can be considered as an estimate of the phoneme accuracy, so the correct label sequence would tend to have the largest $F_2(\mathbf{x}, \mathbf{y}; \theta_2)$ among all possible sequences.
In practice, considering all possible $\mathbf{y}$ is intractable, so only a subset of $\mathbf{Y}$ is considered during training, which will be described later in Subsection~\ref{subsec:train_sdnn}.
\vspace{-3mm}
\subsubsection{Maximizing the margin (Max. Margin)}~\label{subsubsec:cost_margin}
Inspired by the maximum margin concept of structured SVM, we replace the linear part of structured SVM by nonlinear DNN to take advantage of both DNN and maximum margin.
The proposed structured DNN thus optimizes the following formula:
\vspace{-4mm}
\begin{multline}
\min_{\theta_2}{\left \| \theta_2 \right \|^2 + C\sum_{i=1}^{N}{L_i^{\prime}(\theta_2)}}, \\
L_i^{\prime}(\theta_2) = \sum_{\mathbf{y}} max(0, F_2(\mathbf{x}_i, \mathbf{y}; \theta_2) + \Delta(\mathbf{y}_i, \mathbf{y}) \\
- F_2(\mathbf{x}_i, \mathbf{y}_i; \theta_2)) 
\label{equ:sdnn_loss}
\end{multline}
$L_i^{\prime}(\theta_2)$ in (\ref{equ:sdnn_loss}) is parallel to $L_i(\theta_1)$ in (\ref{equ:ssvm_loss}), except that $\theta_1$ and  $F_1(.)$   in (\ref{equ:ssvm_loss}) are replaced by $\theta_2$ and $F_2(.)$ in (\ref{equ:sdnn_loss}) respectively, while the outer max operator in (\ref{equ:ssvm_loss}) is replaced by summation\footnote{We replace the outer max operator in (\ref{equ:ssvm_loss}) of structured SVM with summation. In this way all the label sequences $y$ are properly considered and this makes the DNN training more efficient.}.
Because the loss function $L_i^{\prime}(\theta_2)$ would be larger than zero whenever $F_2(\mathbf{x}_i, \mathbf{y}; \theta_2) + \Delta(\mathbf{y}_i, \mathbf{y}) - F_2(\mathbf{x}_i, \mathbf{y}_i; \theta_2) > 0$, the DNN model parameters $\theta_2$ are penalized if any of the inequalities below do not hold. 
\vspace{-3mm}
\begin{multline}
F_2(\mathbf{x}_i, \mathbf{y}_i; \theta_2) - F_2(\mathbf{x}_i, \mathbf{y}; \theta_2) > \Delta(\mathbf{y}_i, \mathbf{y}), \\ 
i=1, ...., N, \forall{\mathbf{y} \in \mathbf{Y}}.
\end{multline}
Therefore, by (\ref{equ:sdnn_loss}), the scores between the correct label sequence and other label sequences would be separated by at least a margin which is maximized as in structured SVM, but here the scores are evaluated from a DNN with parameters learned based on the DNN framework, rather than from SVM.
Note that all components in loss function $L_i^{\prime}(\theta_2)$ in (\ref{equ:sdnn_loss}) are piecewise differentiable which means we can use back propagation to find the model parameters $\theta_2$ when optimizing (\ref{equ:sdnn_loss}). 
According to (\ref{equ:sdnn_loss}), we need to traverse over all possible $\mathbf{y}$ for an utterance which is intractable and need some approximation as described in Subsection~\ref{subsec:train_sdnn}.

\subsection{Inference with Structured DNN} \label{subsec:test_sdnn}

With the structured DNN trained as above, given the acoustic vector sequence $\mathbf{x}$ of an unknown utterance, we need to find the best phoneme label sequence $\mathbf{y}$ for it. For structured SVM in Subsection~\ref{subsec:ssvm}, due to the linear assumption, the learned model parameter $\theta_1$ contains enough information to execute the Viterbi algorithm to find the best label sequence. This is not true for structured DNN. From (\ref{equ:sl}), in principle we need to search over all possible phoneme label sequences ($K^M$ for $K$ phonemes and $M$ acoustic vectors) for the given acoustic vector sequence and pick the one giving the highest score, which is computationally infeasible.

Instead of searching through all possible phoneme label sequences, we first decode $\mathbf{x}$ using WFST to generate a lattice, and then search through the phoneme label sequences in the lattice which give the highest scores. Obviously, in this way the performance is bounded by the quality of the lattice.

\subsection{Training of Structured DNN} \label{subsec:train_sdnn}

For each training utterance, again we have $K^M$ possible label sequences. It is also impossible to train over all these label sequences for the training utterances. In structured SVM, due to the linear property, we are able to find training examples to produce the maximum margin. For structured DNN here, how to find and choose effective training examples is important. Besides the positive examples (reference phoneme label sequences for the training utterances), in this work negative examples (those other than reference label sequences) are chosen both by random and from the lattice decoded from WFST. For each training utterance with a lattice, the negative examples have three sources: (a) N completely random sequences, (b) N random paths on the lattice, and (c) the N-best paths on the lattice.

\subsection{Full-scale structured DNN (FSDNN)} \label{subsec:fsdnn}

The acoustic feature sequence $\mathbf{x}$ here can be the output of another DNN in the front-end (the DNN in Fig.~\ref{fig:SDNN}(a)).  
For example, it can be generated from a DNN whose input is the filter bank output, and the output is the phoneme posteriorgram vectors.
In this case, during back propagation, we can further propagate the errors of the structured DNN (the DNN in Fig.~\ref{fig:SDNN}(c)) all the way back into the front-end DNN. 
In this way, we have Full-scale Structured DNN (FSDNN) in which all parameters from filter bank up to the whole utterance score are jointly learned.

The FSDNN we proposed can be considered as a special case (or structured version) of Convolutional Neural Network \cite{lawrence1997face} \cite{krizhevsky2012imagenet} which works perfectly in computer vision and speech recognition \cite{lecun1995convolutional}.
The power of CNN is mainly based on shared kernel parameters which are able to discover front-end feature filters. 
In FSDNN, we can view the front-end DNN as the kernel in CNN because they all share the parameters in the front end. 
The difference between CNN and FSDNN is that CNN uses max-pooling layer, while we use $\Psi(\mathbf{x}, \mathbf{y})$ to forward the output of the front-end DNN.

\section{Structured Feature Vector \texorpdfstring{$\Psi(\mathbf{x}, \mathbf{y})$}{} for an utterance}
\label{sec:struct}

Take the filter bank outputs or phoneme posteriorgram vectors as the acoustic vectors for an utterance of $M$ frames, $\mathbf{x} = \{\mathbf{x}^j, j = 1, 2, . . . M \}$, and the phoneme label for $\mathbf{x}^j$ is $\mathbf{y}^j$. So the task is to decode $\mathbf{x}$ into the label sequence $\mathbf{y} = \{y^j, j = 1, 2, ... M \}$. 
Since the most successful and well known solution to this problem is with HMM, we try to encode what HMM has been doing into the feature vector $\Psi(\mathbf{x}, \mathbf{y})$ to be used here.
An HMM consists of a series of states, and two most important sets of parameters -- the transition probabilities between states, and the observation probability distribution for each state. 
Such a structure is slightly complicated for the work here, so in the preliminary work we use a simplified HMM with only one state for each phoneme.
With this simplification, these two sets of probabilistic parameters can be estimated for each utterance by adding up all the counts of the transition between labels (or states) and also adding up all the acoustic vectors for each label (phoneme or state). This is shown in Fig. \ref{fig:psi_ex}(a).

Assume $K$ is the total number of different phonemes, we first define a $K$ dimensional vector $\Lambda(y^j)$ for $y^j$ with its k-th component being 1 and all other components being 0 if $y^j$ is the k-th phoneme. 
Tensor product $\otimes$ is helpful here, which is defined as
\begin{equation}
  \otimes : \mathbb{R}^P \times \mathbb{R}^Q \rightarrow \mathbb{R}^{PQ}, (a\otimes b)_{i+(j-1)P} \equiv a_i \times b_j,
\label{equ:tensor}
\end{equation}
where $\mathbf{a}$ and $\mathbf{b}$ are two ordinary vectors with dimensions $P$ and $Q$ respectively. The right half of (\ref{equ:tensor}) says $\mathbf{a}\otimes \mathbf{b}$ is a vector of dimension $PQ$, whose $\left [ i + (j-1)P \right ]$-th component is the i-th component of $\mathbf{a}$ multiplied by the j-th component of $\mathbf{b}$. With this expression, the feature vector $\Psi(\mathbf{x}, \mathbf{y})$ in Fig. \ref{fig:SDNN}(a) to be used for evaluating the scoring function $F_1(\mathbf{x}, \mathbf{y}; \theta_1)$ in (\ref{equ:ssvm}) or $F_2(\mathbf{x}, \mathbf{y}; \theta_2)$ in (\ref{equ:sdnn}) can then be configured as the concatenation of two vectors,
\begin{equation}
  \label{equ:phi}
  \Psi(\mathbf{x}, \mathbf{y}) = 
  \begin{pmatrix}
    \sum_{j=1}^{M}{\mathbf{x}^j\otimes\Lambda(\mathbf{y}^j)}
    \\ 
    \sum_{j=1}^{M-1}{\Lambda(\mathbf{y}^j)\otimes\Lambda(\mathbf{y}^{j+1})}
  \end{pmatrix},
\end{equation}
where $\mathbf{x} = \{\mathbf{x}^1, \mathbf{x}^2,..., \mathbf{x}^M\}$ and $\mathbf{y} = \{y^1, y^2,..., y^M\}$. The upper half of the right hand side of (\ref{equ:phi}) is to accumulate the distribution of all components of $\mathbf{x}^j$ for each phoneme in the acoustic vector sequence $\mathbf{x}$, and then locate them at different sections of components of the feature vector $\Psi(\mathbf{x}, \mathbf{y})$ (corresponding to the observation probability distribution for each state or phoneme label estimated with the utterance). 
The lower half of the right hand side of (\ref{equ:phi}), on the other hand, is to accumulate the transition counts between each pair of labels (phonemes or states) in the label sequence $\mathbf{y}$ (corresponding to state transition probabilities estimated for the utterance). Then, $\Psi(\mathbf{x}, \mathbf{y})$ is the concatenation of the two, so it keeps the primary statistical parameters of $\mathbf{x}^j$ for different phonemes $y^j$ for all $\mathbf{x}^j$ in $\mathbf{x}$, and the transitions between states for all $y^j$ in $\mathbf{y}$. With enough training utterances $(\mathbf{x}, \mathbf{y})$ and the corresponding function $\Psi(\mathbf{x}, \mathbf{y})$, we can then learn the scoring function $F_1(\mathbf{x}, \mathbf{y}; \theta)$ or $F_2(\mathbf{x}, \mathbf{y}; \theta_2)$ by training the parameters $\theta_1$ or $\theta_2$.
The vector $\Psi(\mathbf{x}, \mathbf{y})$ in (\ref{equ:phi}) can be easily extended to higher order Markov assumptions (transition to the next state depending on more than one previous states). For example, by replacing the upper half of (\ref{equ:phi}) with $\sum_{n=1}^{N}{\mathbf{x}^n\otimes\Lambda(\mathbf{y}^n)\otimes\Lambda(\mathbf{y}^{n+1})}$ and the lower half of (\ref{equ:phi}) with 
$\sum_{n=1}^{N-1}{\Lambda(\mathbf{y}^n)\otimes\Lambda(\mathbf{y}^{n+1})\otimes\Lambda(\mathbf{y}^{n+2})}$
, we have the second order Markov assumption.

Consider a simplified example for $K = 3$ (only 3 allowed phonemes A, B, C) and an utterance with length $M = 4$ as shown in Fig. \ref{fig:psi_ex}(b). It is then easy to find that the upper half of $\Psi(\mathbf{x}, \mathbf{y})$ is $\sum_{n=1}^{4}{\mathbf{x}^n\otimes\Lambda(\mathbf{y}^n)} = {(1.2, 2.6, 2.7, 2.3, 1.5, 2.5 )}' $, and the lower half of $\Psi(\mathbf{x}, \mathbf{y})$ is $\sum_{n=1}^{3}{\Lambda(\mathbf{y}^n)\otimes\Lambda(\mathbf{y}^{n+1})} = {(0,0,0,1,1,0,0,1,0)}'$. We therefore have $\Psi(\mathbf{x}, \mathbf{y}) =\\ {(1.2, 2.6, 2.7, 2.3, 1.5, 2.5, 0, 0, 0, 1, 1, 0, 0, 1, 0)}'$.

\begin{center}
\vspace{-5mm}
\begin{figure}[htb]
  \centering
  \subfloat[a simple example with arbitrary acoustic vector]{{\includegraphics[width=0.17\textwidth]{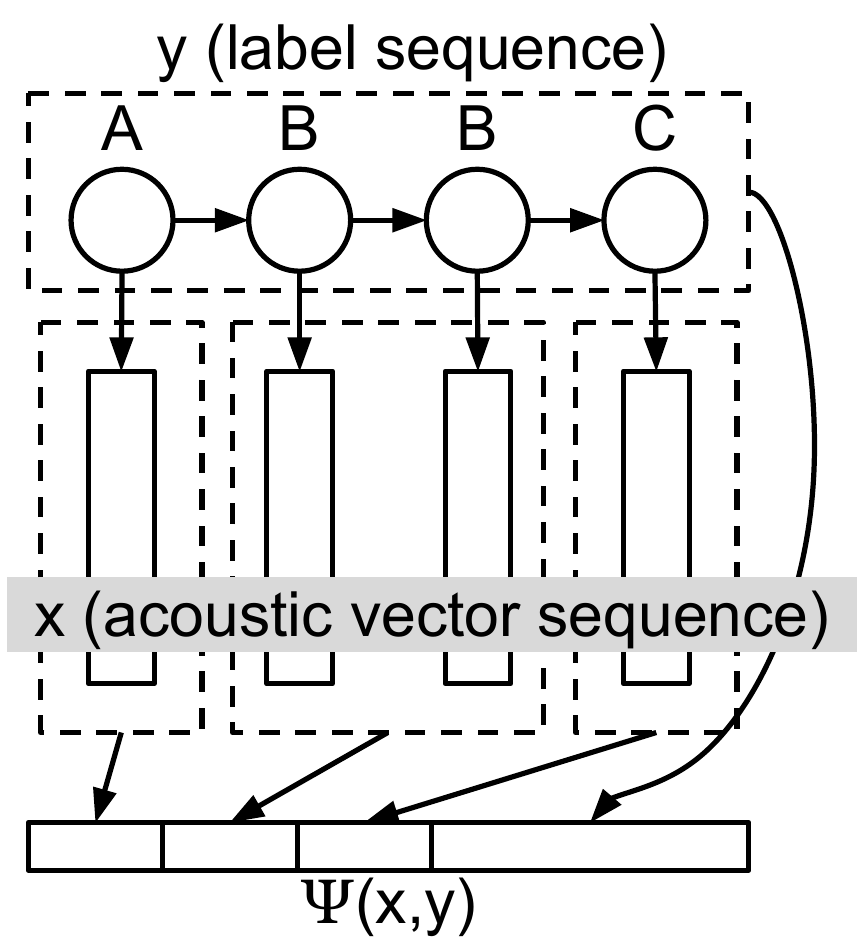} }}%
  \qquad
  \subfloat[a demonstration of how $\Psi(\mathbf{x}, \mathbf{y})$ is computed.]{{\includegraphics[width=0.24\textwidth]{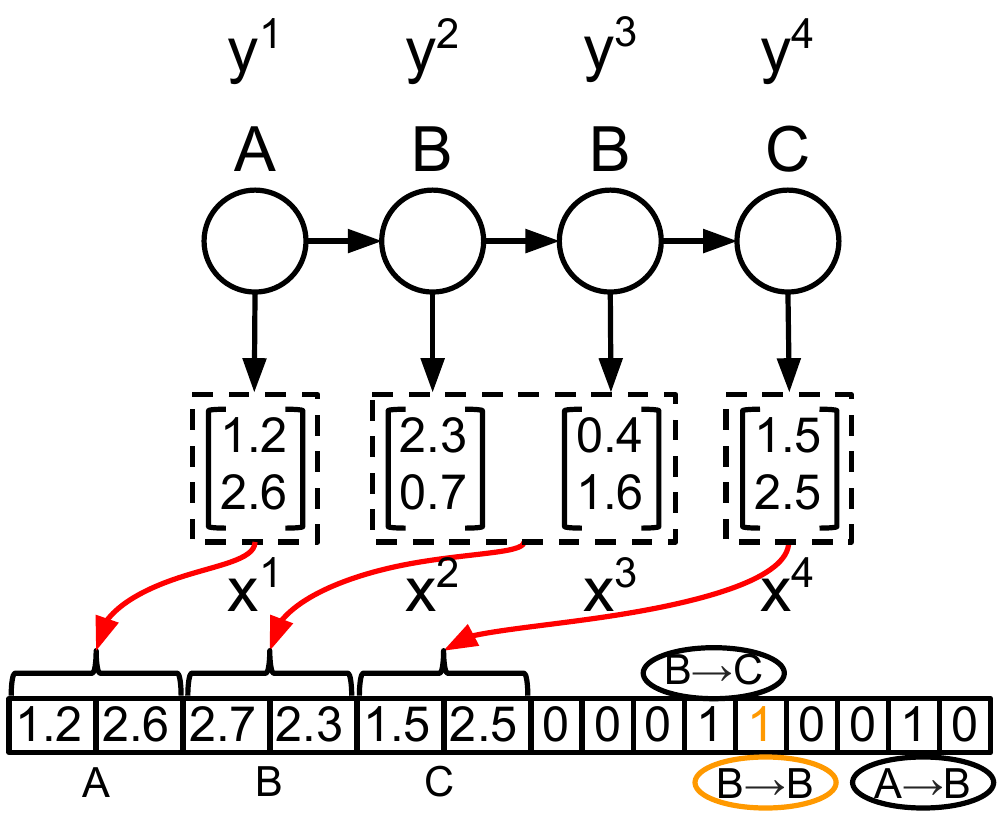} }}%
  \caption{A simplified example of feature sequence $\mathbf{x} = (\mathbf{x}^1, \mathbf{x}^2, \mathbf{x}^3, \mathbf{x}^4 )$ and label sequence $\mathbf{y} = (\mathbf{y}^1, \mathbf{y}^2, \mathbf{y}^3, \mathbf{y}^4 ) = (A, B, B, C) $.}
  \label{fig:psi_ex}
\end{figure}
\end{center}

\vspace{-12mm}
\section{Experimental Setup} \label{sec:setup}
\vspace{-2mm}
Initial experiments were performed with TIMIT. We used the training set without dialect sentences for training and the core testing set (with 24 speakers and no dialect) for testing.
The models were trained with a set of 48 phonemes and tested with a set of 39 phonemes, conformed to CMU/MIT standards\cite{lee1989speaker}.
We used an online library\cite{SSVM_toolkit} for structured SVM, and modified the kaldi\cite{kaldi} code to implement structured DNN. 

Our experiment was based on Vesely's recipe in kaldi, called as baseline, which used LDA-MLLT-fMLLR features obtained from auxiliary GMM models, RBM pre-training, frame cross-entropy training and sMBR. The structured DNN was performed on top of the lattices obtained by Vesely's recipe. We used two sets of acoustic vectors, 
(a)LDA-MLLT-fMLLR feature (40 dimensions), or input to DNN in Vesely's recipe;
(b)phoneme posterior probabilities (48 dimensions) obtained from the 1943 DNN output (state posterior) from Vesely's recipe. Because 1943-dimension feature was too large for $\Psi(\mathbf{x}, \mathbf{y})$, we reduced the dimension to 48(mono-phone size) by adding an extra layer($1943 \times 48$) to Vesely's DNN, and used one-hot mono-phone as training target to train this extra layer.

Unless specified, we used the following parameters: 2 hidden layers, 900 neurons per layer, random initial weights for structured DNN, Vesely's DNN as initial weight for front-end DNN in FSDNN, phone error rate as $\Delta(\cdot,\cdot)$, mini-batch used, momentum = $0.9$, learning rate = $4 \times 10^{-6}$, halving learning rate if the improvements of loss function was too small. Due to computation time, we only used N=1 when choosing the negative examples, that was 1 totally random path, 1-best lattice path and 1 random lattice path were used to train. Test was on $10\times N$ (10-best) lattice paths.


\begin{center}
\begin{table*}
\begin{tabular}{|l||c|c|c|c||c|}
\hline
       Phoneme Error Rate(\%)
     & (A) structured SVM
     & (B) structured DNN
     & (C) structured DNN
     & (D) FSDNN
     & (E) baseline 
     \\ \hline \hline
Loss function   & x  & Approx. Ph. Acc &  max. Margin & max. Margin & x \\ \hline
(1) LDA-MLLT-fMLLR    & 38.62 & 19.98 & 18.03 & \textbf{17.78} & 18.90 \\ 
(2) phoneme posterior & 24.32 & 18.77 & 17.95 & x              & x \\  \hline
\end{tabular}
\caption{
Phoneme Error Rate(\%). 
Rows (1) and (2) are for two different acoustic vector sequence inputs.
Column (A) are the results of structured SVM, while columns (B) and (C) are for structured DNN respectively with the lost function to approximate phoneme accuracy(Approx. Ph. Acc) in Subsection~\ref{subsubsec:cost_mse} and maximizing margin(max. Margin) in Subsection~\ref{subsubsec:cost_margin}. 
Column (D) is for the extension of column (C) into the Full-scale structued DNN (FSDNN) described in Subsection~\ref{subsec:fsdnn}. Column (E) is the baseline (Vesel's recipe).
}
\label{tab:summary}
\vspace{-5mm}
\end{table*}
\end{center}

\vspace{-10mm}
\section{Experimental Results}
\vspace{-2mm}
\label{sec:exp}

The results are listed in Table \ref{tab:summary}. Rows (1) and (2) are for different acoustic features, where row (1) is actually the acoustic features used by Vesel's recipe. Column (A) are the results of structured SVM. Columns (B) and (C) are for the proposed structured DNN, respectively for the lost function of approximating the phoneme accuracy in subsection \ref{subsubsec:cost_mse} in column (B) and maximizing the margin in subsection \ref{subsubsec:cost_margin} in column (C). Column (D) is for the extension of column (C) to the Full-scale structured DNN (FSDNN) described in Subsection~\ref{subsec:fsdnn}, in which the front-end DNN and structured DNN were jointly trained. Column (E) is the baseline. 

It is clear that the phoneme posterior in row (2) is better than the feature vectors used in the Vesel's recipe in row (1), accounting for the powerful feature transform achieved by DNN. The structured DNN outperformed the structured SVM on both acoustic features (columns (B), (C) v.s. (A) on both rows (1) (2)). This is not surprising because the structured SVM learned only the linear transform; while structured DNN, learned much more complex nonlinearity.
Although the results of structured DNN were obtained by rescoring the lattices of Vesel's results which is 18.90\% on Phoneme Error Rate (PER), structured DNN was better than baseline in most cases with both sets of acoustic vectors  (columns (B), (C) v.s. (E)).
These results showed that the proposed structured DNN did learn some substantial information beyond the normal frame-level DNN. 

Comparing the results in columns (B) and (C) of Table~\ref{tab:summary}, we see the selection of the loss function is critical. Margin performed much better than approximating the phoneme accuracy (columns (C) v.s. (B)). A possible reason is as follows. The loss function of approximating the phoneme accuracy may be inevitably dominated by negative examples, since we used much more negative examples than positive examples in training, and the positive and negative examples were equally weighted. The loss function of maximizing the margin, on the other hand, focused on the score difference between each pair of positive and negative examples, and as a result, the positive and negative examples were weighted equally even if very different numbers. 

When we compare the structured DNN with the full-scale structured DNN (FSDNN) using the same loss function (columns (D) v.s. (C)), we see that propagating the errors all the way back into the front-end layer did offer good improvement, and the best result we got here is 17.78\%, which beat baseline 18.90\% by an 5.5\% relative improvements (columns (D) v.s. (E) in row (1)).
Note that the full potential of structured DNN was not well explored yey, for example, as explained in Section \ref{sec:struct}, we simply assume a single state for a phoneme in (\ref{equ:phi}), which is certainly over-simplified. Also, as mentioned in Sections \ref{subsec:train_sdnn}, \ref{subsec:test_sdnn}, both the inference and training were simplified for reduced computation and lack of time.

\begin{figure}[htb]
\begin{minipage}[b]{.48\linewidth}
  \centering
  \centerline{\includegraphics[width=4cm]{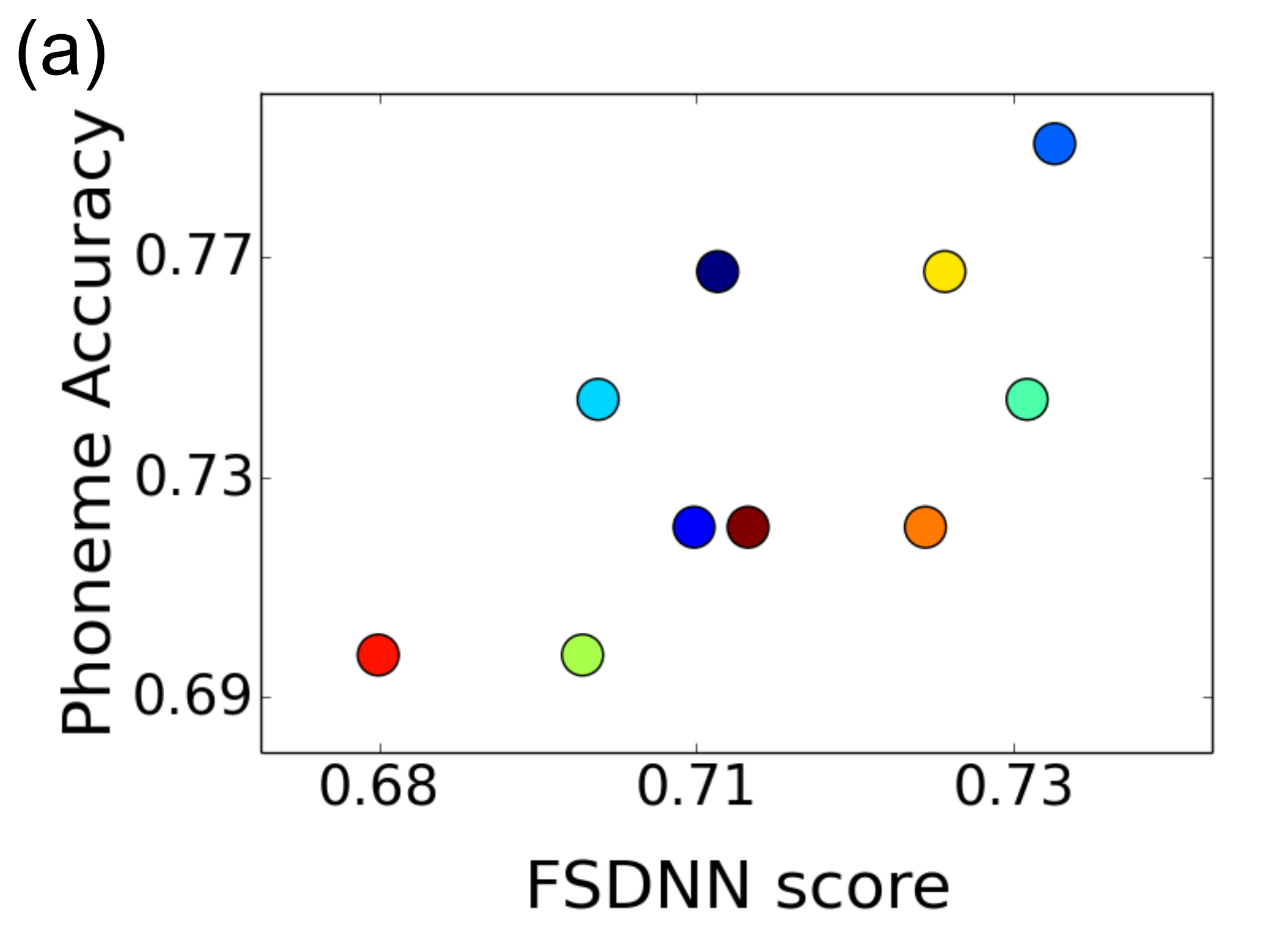}}
\end{minipage}
\begin{minipage}[b]{.48\linewidth}
  \centering
  \centerline{\includegraphics[width=4cm]{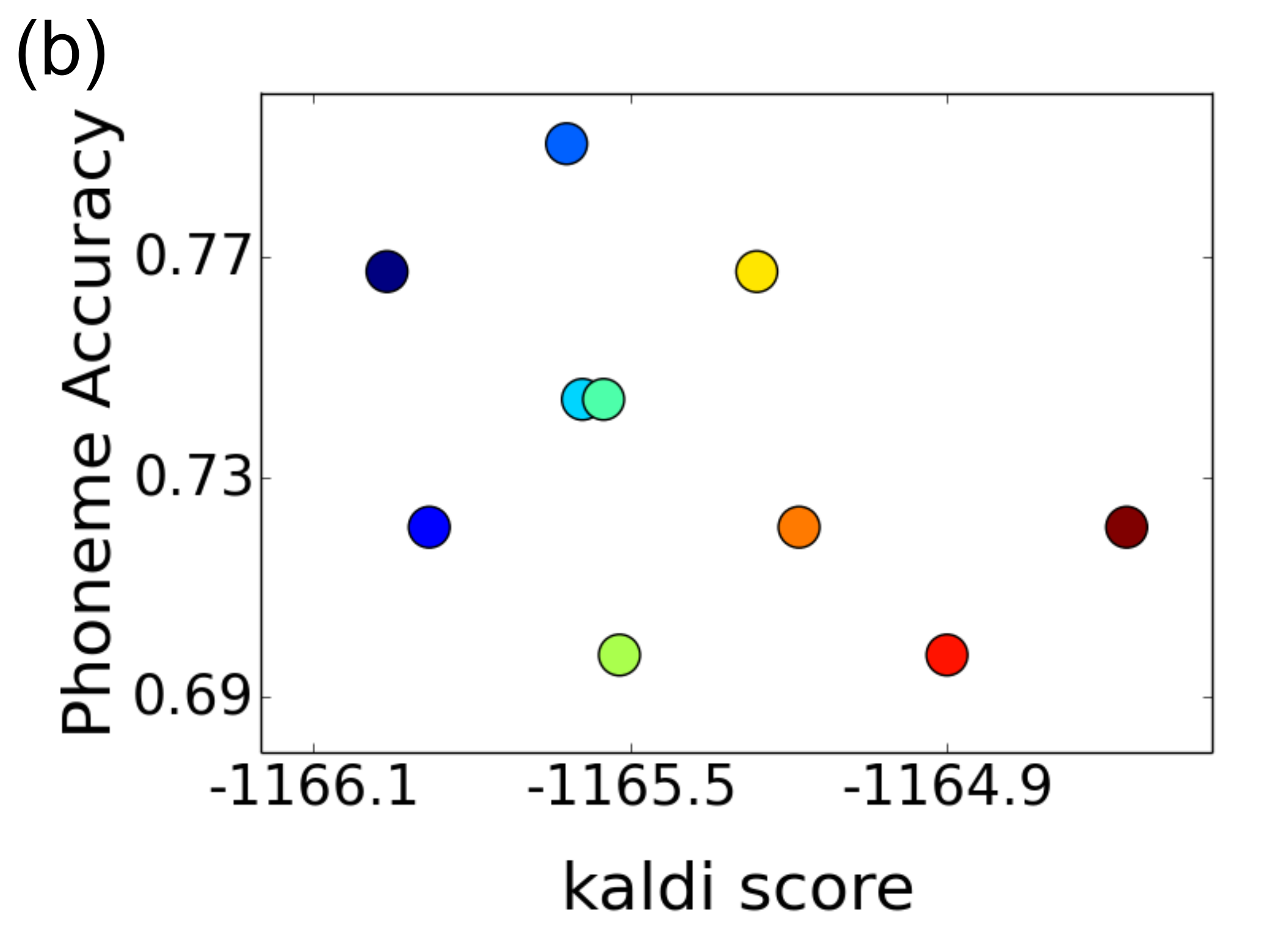}}
\end{minipage}
  \caption{ The phoneme accuracy (vertical scale) evaluate for the 10-best paths for a selected example utterance plotted as a function of the scores obtained in the recognizer (horizontal scale): (a) FSDNN score in FSDNN and (b) kaldi score. There are 10 dots in each figure, each for a path among the best 10. The same color was used to mark the same path in (a) and (b). }
  \vspace{-5mm}
  \label{fig:score_ex}
\end{figure}

In order to see why FSDNN in column (D) can do better than baseline in column (E), we took a deeper look at the data, and the results of a selected example utterance was shown in Figure \ref{fig:score_ex}. In this figure, the phoneme accuracy (vertical scale) for the 10-best paths of the example utterance was plotted as a function of the scores obtained in the recognizer (horizontal scale), i.e., FSDNN and baseline in columns (D)(E) and row (1). There are 10 dots in each figure, each for a path among the best 10. The same color was used to mark the same path.
Note that the phoneme accuracy is discrete here for a single utterance of the integer number for the phoneme errors. In Figure \ref{fig:score_ex}(a), the FSDNN gave the highest score to the path with the highest phoneme accuracy (the upper right blue point). In Figure \ref{fig:score_ex}(b), however, this blue point received only relatively low score from kaldi (top of the figure), while a path with lower phoneme accuracy had the highest score from kaldi (the right most brown point). This resulted in a lower phoneme accuracy by baseline for this utterance. When we evaluated the regression line for those figures, we found FSDNN score and phoneme accuracy are positively correlated in Figure \ref{fig:score_ex}(a), but the kaldi score and phoneme accuracy are negatively correlated. 
Although this is for just a selected example, it is easy to find many such examples with similar situation. Noting that FSDNN here was trained on the large margin criteria not considering the phoneme accuracy, but what was learned was positively correlated with the phonme accuracy.

\begin{figure}[htb]
  \centering
  \includegraphics[width=0.8\linewidth]{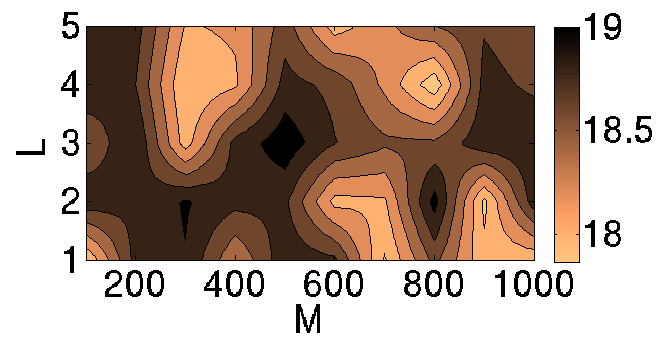}
  \vspace{-3mm}
  \caption{{Phoneme Error Rate(\%) map for the proposed full-scale structured DNN (FSDNN) for different values of $L$ (number of hidden layers) and $M$ (number of neurons per layer).}}
  \label{fig:dnn}
\end{figure}

The next experiment is to analyze the PER for different choices of the key hyper-parameters for the full-scale structured DNN (FSDNN), number of hidden layers $L$ and number of neurons $M$ in each hidden layer.
Figure \ref{fig:dnn} is the result, a visualized PER map for FSDNN using acoustic vectors (1).  The horizontal axis is $M$ where $M = 100, 200, ... 1000$, and the vertical axis is $L$ where $L = 1, 2, ... 5$. Therefore, the figure consists of $5\times 10 = 50$ data points. The overall performance is approximately between 17\% and 19\%, more or less comparable to baseline. 
For this task, $N=1$ for training, $N=10$ for inferencing. Better PER seemed to be located at several disjoint regions (4 valleys in the figure). The best result is on $(L, M) = (4, 800)$, which was the case in Table \ref{tab:summary}. The disjoint valleys may come from the relatively poor learning due to lack of training data and poor initialization. The loss function in (\ref{equ:sdnn_loss}) is ReLU like, which may result in a similar behavior of ReLU training (highly dependent on the initialization). 
\vspace{-3mm}

\vspace{-8mm}
\section{Conclusion and Future Work}
\vspace{-3mm}
\label{sec:con}
In this paper, we propose a new structured learning architecture, structured DNN, for phoneme recognition which jointly considers the structures of acoustic vector sequences and phoneme label sequences globally. Preliminary test results show that the structured DNN outperformed the previously proposed structured SVM and beat the state-of-the-art kaldi results. We will work on multiple states per phoneme in the future, and explore more possibilities of this approach.

\newpage
\bibliographystyle{IEEEbib}
\bibliography{mybib}

\end{document}